\documentclass[10pt,twocolumn,letterpaper]{article}

 \usepackage{cvpr}              
\usepackage{booktabs}
\usepackage{adjustbox}   
\usepackage{caption}
\usepackage{pifont}
\usepackage[dvipsnames,svgnames]{xcolor}

\usepackage{bm} 
\newcommand{\cmark}{\ding{51}} 
\newcommand{\xmark}{\ding{55}} 

\input{cvpr_macros}

\definecolor{cvprblue}{rgb}{0.21,0.49,0.74}
\usepackage[pagebackref,breaklinks,colorlinks,allcolors=cvprblue]{hyperref}

\def\paperID{9} 
\def\confName{CVPR}
\def\confYear{2026}

\title{When Negation Is a Geometry Problem in Vision-Language Models}

\author{
Fawaz Sammani$^{1,2,*}$, Tzoulio Chamiti$^{1,2,*}$, Paul Gavrikov$^{3}$, Nikos Deligiannis$^{1,2}$ \\
$^{1}$ ETRO Department, Vrije Universiteit Brussel, Belgium \\
$^{2}$ imec, Kapeldreef 75, B-3001 Leuven, Belgium \quad
$^{3}$ Independent Researcher \\
$^{*}$ Equal contribution
}

\begin{document}
\maketitle

\begin{abstract}
Joint Vision-Language Embedding models such as CLIP typically fail at understanding negation in text queries—for example, failing to distinguish ``no" in the query: ``a plain blue shirt with no logos". Prior work has largely addressed this limitation through data-centric approaches, fine-tuning CLIP on large-scale synthetic negation datasets. However, these efforts are commonly evaluated using retrieval-based metrics that cannot reliably reflect whether negation is actually understood. In this paper, we identify two key limitations of such evaluation metrics and investigate an alternative evaluation framework based on Multimodal LLMs-as-a-judge, which typically excel at understanding simple yes/no questions about image content, providing a fair evaluation of negation understanding in CLIP models. We then ask whether there already exists a direction in the CLIP embedding space associated with negation. We find evidence that such a direction exists, and show that it can be manipulated through test-time intervention via representation engineering to steer CLIP toward negation-aware behavior without any fine-tuning. Finally, we test negation understanding on non-common image-text samples to evaluate generalization under distribution shifts\footnote{https://github.com/fawazsammani/negation-steering}.
\end{abstract}

\section{Introduction}
\label{sec:intro}
\begin{figure}
    \centering
    \includegraphics[width=\linewidth]{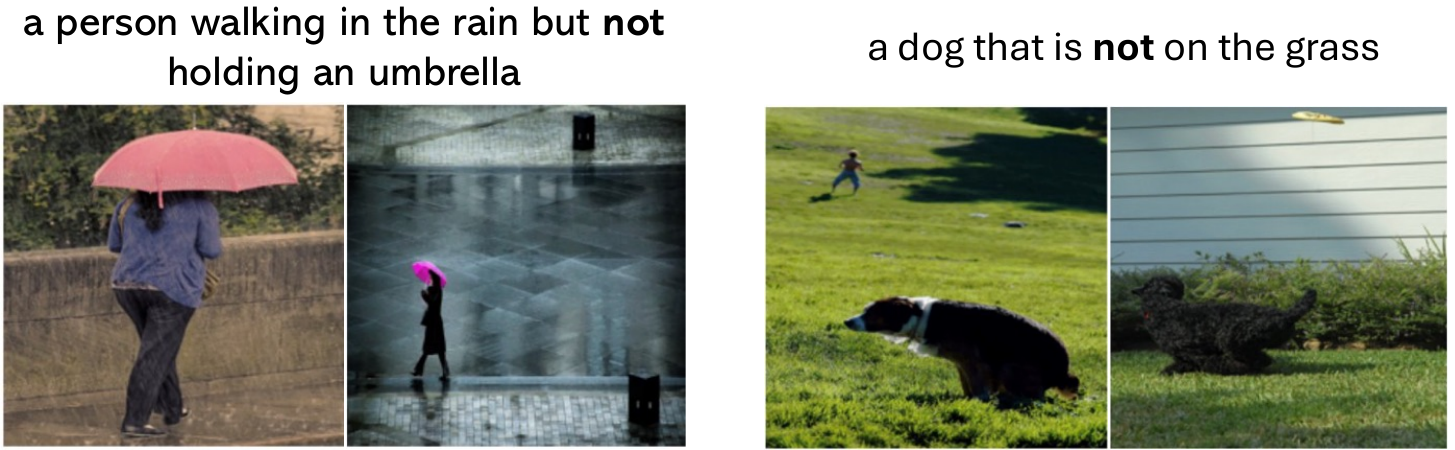}
    \caption{CLIP models fail at understanding negated text queries, retrieving images that disregard negation cues.}
    \label{fig:demo}
\end{figure}

Contrastive Language–Image Pretraining (CLIP) \cite{Radford2021LearningTV} learns a joint embedding space that aligns images and text, where cosine similarity between their embeddings serves as a measure of semantic correspondence, with higher similarity indicating a stronger match. Despite their effectiveness, CLIP models struggle with understanding negation (\cref{fig:demo}), an equally critical aspect of language. Many real-world applications require robust negation handling, particularly in search and recommendation systems where CLIP models often serve as the primary building block. For instance, a user may search for images of “lion sightings with \textit{no} vehicles in frame”, a city planner for “high-traffic intersections with \textit{no} pedestrian crossings”, or a real-estate agent for “apartments with a balcony and \textit{no} busy market street”. In each case, a single negation cue fundamentally alters the expected result. This limitation impacts a wide range of multimodal applications built on CLIP, including text-to-image generation \cite{Rombach2021HighResolutionIS, podell2024sdxl}, multimodal large language models \cite{Liu2023ImprovedBW, li2025llavaonevision, Li2023OtterAM}, open-vocabulary object detection \cite{minderer2023scaling, zhang2025cyclic}, referring image segmentation \cite{Lee2023WeaklySR, Lee2024TowardIR, Lddecke2021ImageSU}, and more.  We discuss reasons for this phenomenon in \cref{relwork}. 

\begin{figure*}[ht]
    \centering
    \includegraphics[width=1\linewidth]{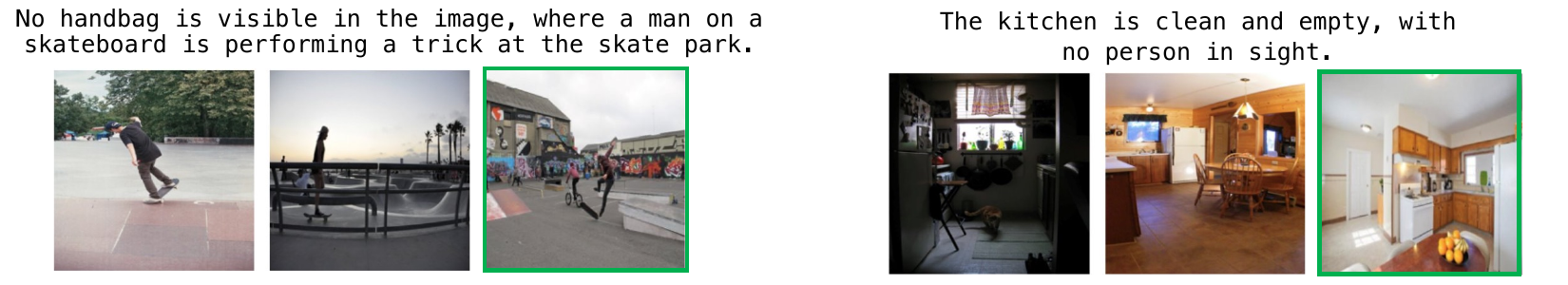}
    \caption{Negated Text-to-Image Retrieval synthetic benchmarks introduce False Negatives, images retrieved that correctly match the query (first two images) but are not labeled as the ground-truth image (last image highlighted in green). Examples from NegBench \cite{Alhamoud2025VisionLanguageMD}}
    \label{fig:plausability}
\end{figure*}

To counteract this issue, existing approaches adopt data-centric strategies, generating synthetic negated captions either through template-based methods \cite{Singh2025LearningTP} or by leveraging a range of foundation models \cite{yuksekgonul2023when, Alhamoud2025VisionLanguageMD, kang2025clip, Park2025KnowB}. While scalable and efficient, this creates a massive amount of false negatives (\ie, images satisfying the query but not explicitly labeled as ground truth by the data construction pipeline). This issue is particularly relevant when evaluating with the commonly used COCO dataset \cite{Lin2014MicrosoftCC}, where key themes are often similar. We further observe that some baseline models collapse after being finetuned and result in retrieved images that are unrelated to the query. This motivates us to investigate alternative methods for the evaluation of retrieval and negation understanding using MLLM-as-a-judge, separately evaluating each component of the query and the retrieved image. Since this approach is unsupervised, it can be applied to any image database of any size.

Finally, we investigate whether a ``negation" direction vector already exists within the CLIP embedding space. Specifically, we ask whether CLIP embeddings (or a subset of their dimensions) encode information related to negation, and if so, whether the embedding space can be manipulated at inference time by steering it towards the discovered ``negation" direction when a query contains negation. In our work, we find evidence of this directional vector in the CLIP space, and we show that CLIP models can be geometrically steered to understand negation without any finetuning. 

\noindent To summarize, our main contributions are:
\begin{itemize}
    \item We highlight two fundamental limitations in the evaluation of negated retrieval systems, and explore alternative evaluation methods based on MLLM-as-a-judge.
    \item We investigate whether a negation direction already exists in the CLIP embedding space and whether it can be extracted. 
    \item We find evidence that this direction already exists, and we use it to steer CLIP representations at inference-time for negation-aware understanding.
\end{itemize}

\section{Related Work}
\label{relwork}
\textbf{Negation in Vision-Language-Models:} Several works \cite{yuksekgonul2023when, Alhamoud2025VisionLanguageMD, kang2025clip, Park2025KnowB} have investigated why joint embedding VLMs such as CLIP typically collapse affirmative (\eg, a dog on the grass) and negated (\eg, a dog not on the grass) queries into indistinguishable embeddings. The reasons are two-fold: \textbf{(1)} CLIP behaves like a bag of words \cite{yuksekgonul2023when, koishigarina2026clip}, primarily contextualizing the prompt as a whole rather than understanding fine-grained linguistic structures, and relying heavily on the presence of content words rather than the syntactic and relational structure of the sentence. This limitation largely stems from the nature of CLIP’s pretraining data: captions involving negation are severely underrepresented \cite{Park2025KnowB}. For instance, it is far less common to encounter an internet caption of ``a dog not on the grass". Even when such captions appear in the training data, contrastive learning—despite very large batch sizes—rarely provides paired positive and negative samples in the same batch that share the same content but differ in relational structure or word order. As a result, models seldom observe contrasts (\eg, ``a dog on the grass" vs. ``a dog not on the grass"). A bag-of-words strategy can be high-reward and sufficient to get a low loss. \textbf{(2)}~In addition to the training data, the geometry of a single embedding vector from CLIP, one for an image and one for a text, makes it unreliable at handling basic matching, attribute binding, spatial relationships, and negation all together. This problem occurs even in large and better-performing CLIP models such as CLIP ViT-L/H \cite{yuksekgonul2023when}, SigLIP ViT-L/14 \cite{zhai2023SigmoidLF}. 

\noindent \textbf{Data-Centric Approaches to Negation.}
To address these limitations, existing approaches adopt data-centric strategies of generating synthetic negated captions and finetuning CLIP on them \cite{Singh2025LearningTP, Alhamoud2025VisionLanguageMD, kang2025clip}. While this strategy enables large-scale training, the synthetic data generation process can introduce a substantial number of false negatives, where multiple images may satisfy the same negated caption, but only a single image is treated as a positive example. As a result, improvements measured under standard retrieval-based metrics mix semantic matching with genuine negation understanding. Apart from this, we also show that some baseline models collapse when being finetuned on synthetic negation data, leading to unrelated image associations to the text query. 

These observations motivate us to investigate an alternative evaluation metric based on MLLM-as-a-judge, which jointly assesses retrieval and negation understanding. Moreover, in contrast to prior approaches that rely on fine-tuning CLIP with large volumes of synthetic data, we instead examine whether a latent ``negation" direction already exists within CLIP’s embedding space, and whether CLIP can be manipulated with this direction via representation engineering to improve its handling of negation.


\section{Negation Evaluation in Image Retrieval}
\label{sec:metric-llm-as-a-judge}
All existing approaches to negation-aware vision–language learning \cite{yuksekgonul2023when, Singh2025LearningTP, Alhamoud2025VisionLanguageMD, kang2025clip, Park2025KnowB} rely on the same underlying strategy: they generate synthetic negated captions and fine-tune CLIP on this data. For example, NegBench \cite{Alhamoud2025VisionLanguageMD} constructs a large-scale synthetic dataset of 12 million image–text pairs by combining the OWL-ViT open-vocabulary object detection model \cite{Minderer2022SimpleOO} with a Large Language Model (LLM). While this pipeline is scalable and efficient, it introduces two fundamental problems:

\noindent \textbf{(1) False Negatives:} The automatic data annotation pipeline introduces a substantial amount of false negatives. These are images that are correct matches for a caption, but are not labeled as ground truth by the automatic process. The COCO dataset contains many similar images with overlapping object compositions. Since the annotation pipeline operates in isolation per image–caption pair, it is highly likely that multiple images satisfy the same (negated) caption. When only a single image is treated as the positive example, all other valid matches are implicitly considered incorrect during evaluation. We provide an example of this phenomenon in \cref{fig:plausability}, where the ground-truth labeled image is outlined in green. Notably, this problem is also well-studied in conventional image-text retrieval \cite{chun2022eccv_caption}, motivating researchers to re-annotate data. 

\begin{figure}
    \centering
    \includegraphics[width=1\linewidth]{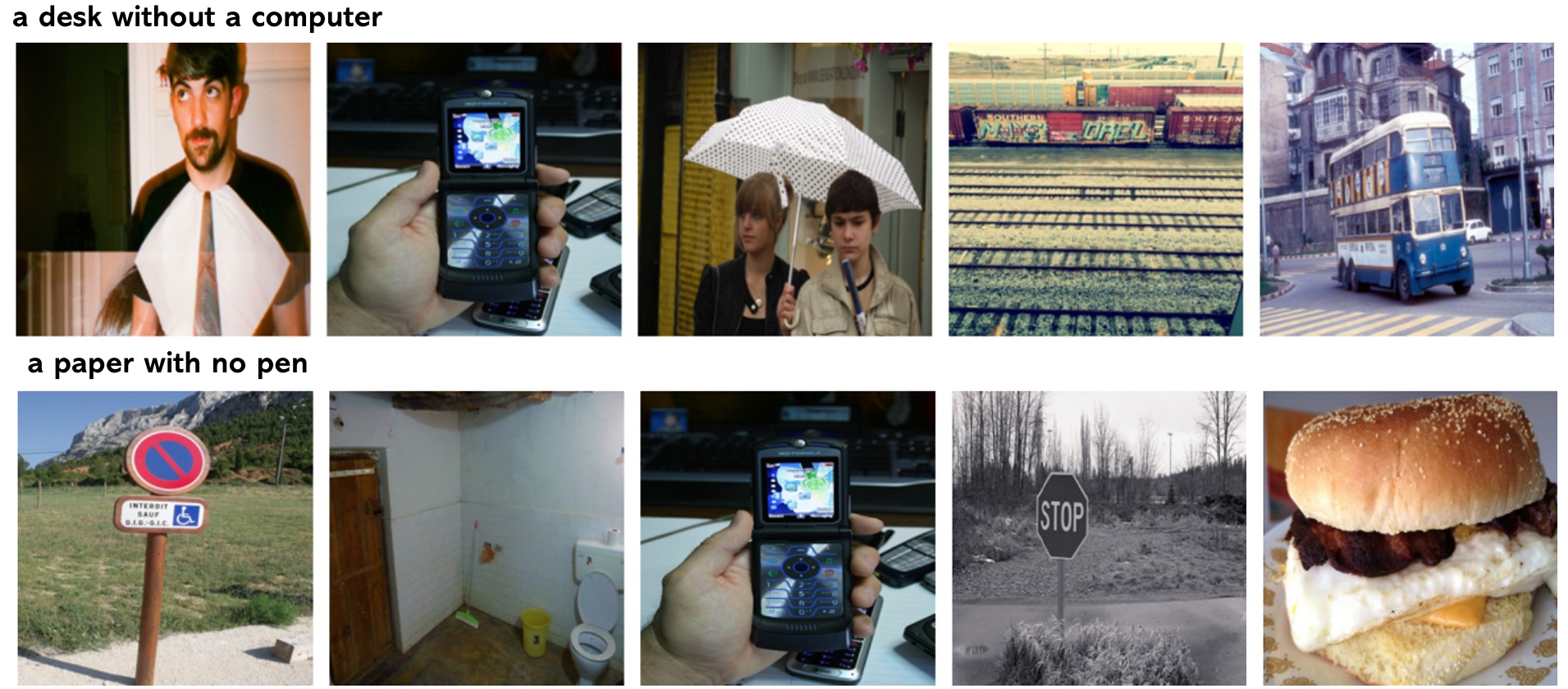}
    \caption{ConCLIP results in collapse after finetuning, retrieving completely irrelevant images to the text query.}
    \label{fig:conclip_ood}
\end{figure}

\noindent \textbf{(2) Collapse After Finetuning:} We also observe that ConCLIP \cite{Singh2025LearningTP}, one of the baseline models, collapses completely when handling negated queries. Specifically, it retrieves images that are entirely unrelated to the negated text query and assigns them lower cosine similarity scores than any non-negated query, as illustrated in \cref{fig:conclip_ood}. This finding is consistent with \cite{Alhamoud2025VisionLanguageMD}, which observed that ConCLIP collapses \textit{all} negated captions (independent of the object or action being negated) into the same point in the embedding space. Such behavior also raises concerns about the reliability of the evaluation metric used in \cite{Singh2025LearningTP}. 

\noindent Motivated by these two observations, we explore alternative evaluation strategies for assessing retrieval and negation understanding. Multimodal Large Language Models (MLLMs) achieve remarkable performance on complex visual reasoning benchmarks. This implies that they can reliably judge simpler yes/no questions about image content. Drawing inspiration from the success of LLM-as-a-judge~\cite{gu2024survey}, we propose to use MLLM-as-a-judge to evaluate text-to-image negated retrieval performance. This also allows us to use \textit{any} image database of \textit{any} size, simulating real-world scenarios where database sizes may be prohibitively large. Unlike existing negation benchmarks, we use a relatively large image database of 25K images (5$\times$ larger than current benchmarks). 

\begin{figure}
    \centering
    \includegraphics[width=1\linewidth]{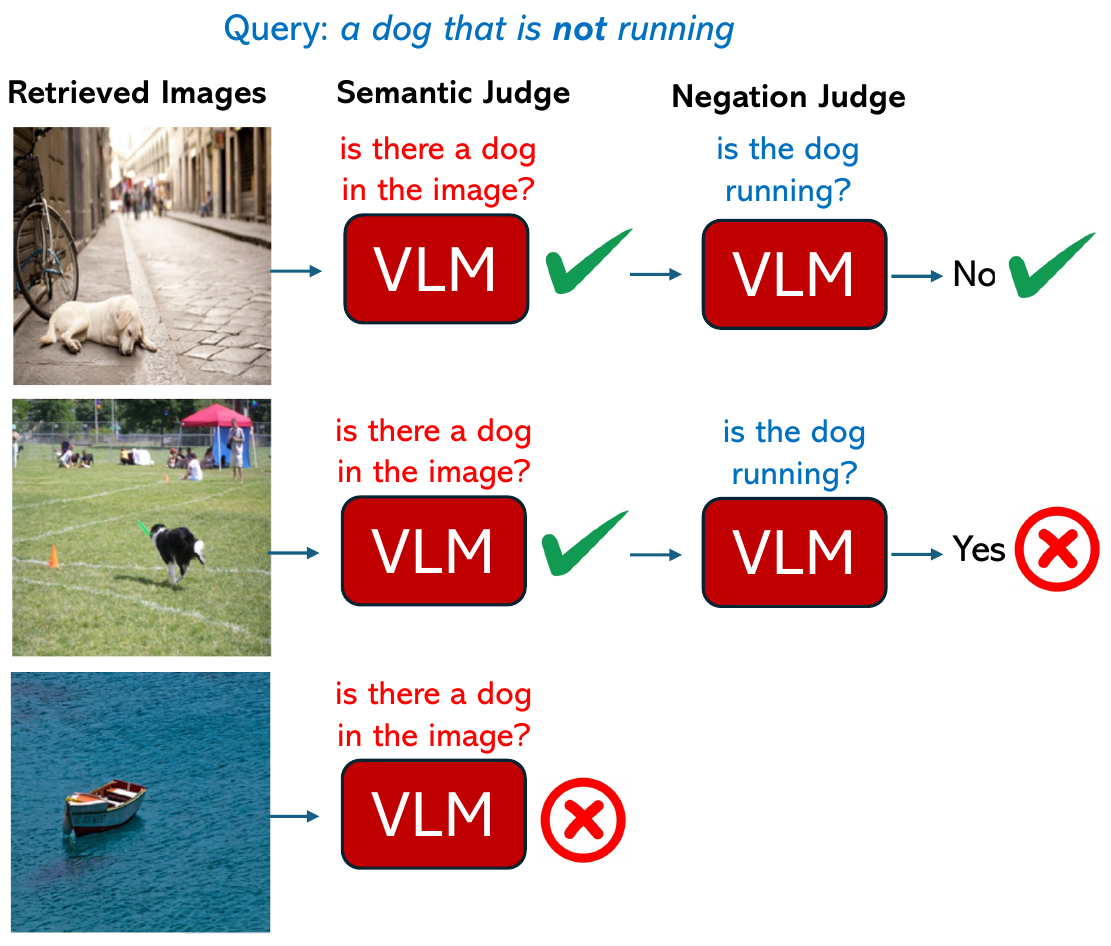}
    \caption{Our MLLM-as-a-judge evaluation framework for retrieval and negation. The same MLLM is used to assess both tasks.}
    \label{fig:llm-as-as-judge}
\end{figure}

\noindent \textbf{MLLM-as-a-judge evaluation protocol:} Let $c_n$ denote a text query that contains negation. We use an LLM to formulate two evaluation questions, $q_r$ and $q_n$. The question $q_r$ assesses whether the retrieved image is contextually and semantically correct, mitigating the collapse issue discussed earlier. The question $q_n$ evaluates negation (\ie, whether the negated object is present in the image). Evaluation proceeds sequentially. We first ask $q_r$, whose ground-truth answer is ``yes'', indicating that the retrieved image is contextually and semantically correct. If the judge answers ``yes'' (\ie, $q_r$ is correct), we then ask $q_n$, whose ground-truth answer is ``no'', indicating that the negated object is not present in the image. For negation benchmarks, both $q_r$ and $q_n$ must be answered correctly. \cref{fig:llm-as-as-judge} illustrates three example outcomes: (i) both questions answered correctly, (ii) only the semantic and contextual correctness ($q_r$) answered correctly, and (iii) neither question answered correctly. Our evaluation pipeline is shown in \cref{fig:llm-as-as-judge}. In our work, we use Qwen-3-VL-32B, Qwen-3-VL-8B and Qwen-3-VL-4B \cite{Qwen3-VL} as MLLM judges. At the time of writing, Qwen-3-VL is regarded as one of the strongest open-source MLLMs, which motivates our choice.  

\begin{table*}
\centering
\caption{Retrieval and Negation performance of baseline models with \textbf{ViT-B/32} on NegBench across two MLLMs-as-a-judge.}
\setlength{\tabcolsep}{5.5pt}
\begin{tabular}{lccc|ccc|ccc|ccc}
\toprule
 & \multicolumn{6}{c|}{\textbf{Qwen3-VL-8B}} & \multicolumn{6}{c}{\textbf{Qwen3-VL-32B}} \\
\cmidrule(lr){2-7}\cmidrule(lr){8-13}
 & \multicolumn{3}{c}{Retrieval} & \multicolumn{3}{c|}{Retrieval and Negation}
 & \multicolumn{3}{c}{Retrieval} & \multicolumn{3}{c}{Retrieval and Negation} \\
\cmidrule(lr){2-4}\cmidrule(lr){5-7}\cmidrule(lr){8-10}\cmidrule(lr){11-13}
Model & Top-1 & Avg.5 & Top-5 & Top-1 & Avg.5 & Top-5 & Top-1 & Avg.5 & Top-5 & Top-1 & Avg.5 & Top-5 \\
\midrule
CLIP \cite{Radford2021LearningTV}      & 46.7 & 30.4 & 75.2 & 39.4 & 25.1 & 68.4 & 47.2 & 30.6 & 76.3 & 40.3 & 25.4 & 69.9 \\
ConCLIP \cite{Singh2025LearningTP}    & 46.6 & 30.7 & 76.4 & 40.8 & 26.1 & 70.7 & 47.2 & 30.7 & 77.7 & 41.6 & 26.3 & 72.2 \\
NegCLIP \cite{yuksekgonul2023when}    & \textbf{60.9} & \textbf{37.9} & \textbf{84.3} & \textbf{51.7} & \textbf{31.5} & \textbf{77.9} & \textbf{62.0} & \textbf{38.3} & \textbf{85.8} & \textbf{53.1} & \textbf{32.1} & \textbf{79.8} \\
CLIP-CC12M \cite{Alhamoud2025VisionLanguageMD} & 49.0 & 31.2 & 76.4 & 42.7 & 26.7 & 70.6 & 49.9 & 31.4 & 78.0 & 43.8 & 27.0 & 72.5 \\
\bottomrule
\end{tabular}
\label{tab:retrieval_and_negation_negbench}
\end{table*}

\noindent \textbf{Results on NegBench:} We report results on the NegBench benchmark \cite{Alhamoud2025VisionLanguageMD}. NegBench consists of a database of 5K images from the COCO 2017 validation set \cite{Lin2014MicrosoftCC}, each annotated with 5 synthetic negated captions. Results are shown in \cref{tab:retrieval_and_negation_negbench} using CLIP ViT-B/32. We compared the original CLIP to three baselines in the literature: ConCLIP \cite{Singh2025LearningTP}, NegCLIP \cite{yuksekgonul2023when}, and CLIP-CC12M \cite{Alhamoud2025VisionLanguageMD}. We focus on the ViT-B/32 backbone because both NegCLIP and CLIP-CC12M are only available for CLIP ViT-B/32. We report three metrics: Top-1, Avg.5, and Top-5, both for retrieval performance (correctness of $q_r$) and for joint retrieval and negation performance (correctness of $q_r$ \textit{and} $q_n$). Top-1 measures whether the highest-ranked retrieved image is correct. Avg.5 is the mean score over the top-5 retrieved images. Top-5 measures whether at least one of the top-5 retrieved images is correct. As shown, our evaluation protocol indicates that all baselines, except NegCLIP, perform only marginally better than the original CLIP. NegCLIP \cite{yuksekgonul2023when} is the only method that significantly outperforms all other baselines. By analyzing the training data of both NegCLIP and CLIP-CC12M, we find that NegCLIP was trained on contrasting data covering (1) attribute binding (\eg, a child kissing an old person \textit{vs} an old person kissing a child), (2) spatial relationships (\eg, A left of B \textit{vs} B left of A), and (3) negation. On the other hand, the training data of CLIP-CC12M is largely dominated by negation data. This explains why NegCLIP outperforms CLIP-CC12M.

\begin{table}[h]
\centering
\caption{Retrieval performance (Recall@K) of baseline models with \textbf{ViT-B/32} on NegBench.}
\setlength{\tabcolsep}{6pt}
\begin{tabular}{lccc}
\toprule
Model & R@1 & R@5 & R@10 \\
\midrule
CLIP \cite{Radford2021LearningTV}            & 0.24 & 0.47 & 0.59 \\
ConCLIP \cite{Singh2025LearningTP}          & 0.26 & 0.51 & 0.62 \\
NegCLIP \cite{yuksekgonul2023when}          & \textbf{0.37} & \textbf{0.64} & \textbf{0.74} \\
CLIP-CC12M \cite{Alhamoud2025VisionLanguageMD} & 0.28 & 0.51 & 0.62 \\
\bottomrule
\end{tabular}
\label{tab:retrieval_negbench}
\end{table}

\noindent We also report standard retrieval performance on NegBench using Recall@K in \cref{tab:retrieval_negbench}. Recall@K measures whether at least one of the top-$K$ retrieved images is labeled as a correct match under the benchmark annotations. Under these metrics, all negation-aware models appear to improve smoothly over the CLIP baseline,
with NegCLIP achieving the highest recall. However, such an evaluation does not distinguish between semantic retrieval correctness and negation satisfaction, nor does it account for the presence of multiple valid matches that are not annotated as ground truth. As shown in \cref{fig:plausability}, retrieved images can be semantically correct and satisfy the negated caption while still being counted as incorrect due to false negatives in the benchmark. On the other hand, MLLM-as-a-judge evaluation alleviates this problem.

\section{A Negation Direction in the CLIP space}
\label{sec:negation_dir}
Although current works address the negation issue using data-centric approaches, an earlier work \cite{quantmeyer-etal-2024-clip} observed that CLIP already possesses an inherent ability to process negation. This observation led us to hypothesize that a directional vector for negation may already exist within the CLIP embedding space. In this work, we draw inspiration from TCAV \cite{Kim2017InterpretabilityBF}, an established work in the field of Explainable AI, to verify this hypothesis. 

Specifically, we take 4,000 captions from the COCO dataset \cite{Lin2014MicrosoftCC} and use an LLM to negate them. Thus, we obtain two subsets of the original captions and their negated counterparts, each of equal size. We also instruct the LLM to vary the negation cues to avoid overfitting to a single style. We extract the hidden representations of all captions and their negations from the residual stream of a layer $l$ of the CLIP text encoder. Specifically, we use the hidden state corresponding to the \texttt{<eos>} token, denoted as $h^l \in R^d$, as it serves as a summary representation of the entire caption. This choice is further backed by \cite{quantmeyer-etal-2024-clip}, which demonstrates that the final token position has the strongest influence on the negation signal. We split the dataset into training and test sets and train a linear binary classifier to distinguish between original captions (label 0, no negation) and their negated counterparts (label 1, negated). 

\begin{figure}
    \centering
    \includegraphics[width=\linewidth]{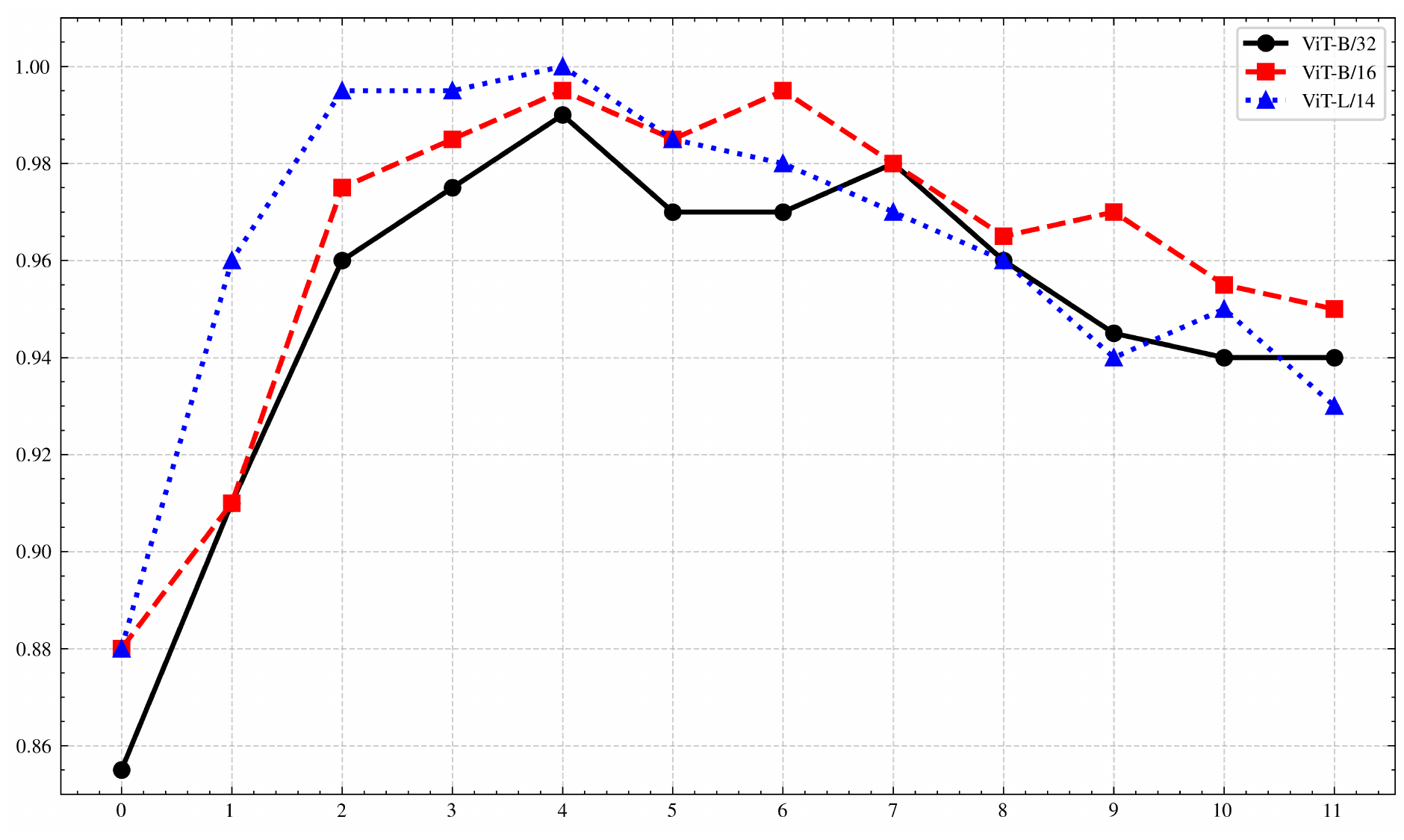}
    \caption{Linear classifier performance to distinguish between a caption and its negation. x-axis represents the layer number, and y-axis represents the test accuracy.}
    \label{fig:classifier}
\end{figure}

The classifier’s performance is shown in \cref{fig:classifier} for 3 CLIP models: ViT-B/32, ViT-B/16, and ViT-L/14. As shown in \cref{fig:classifier}, all CLIP models achieve a test accuracy of 99\% or above at layer 4, indicating that negation-related information is clearly encoded in the hidden states and that the representations of original and negated captions are linearly separable. If negation information were not encoded—such that both types of captions collapsed to similar representations—the classifier would not be able to reliably distinguish between them. We further observe that negation information is better encoded in intermediate layers of the CLIP Text Encoder, rather than early or late layers. Building on these two observations, we investigate whether we can manipulate the CLIP text encoder embeddings using this directional vector to steer them toward the direction of negation.

\begin{table*}
\centering
\caption{Intervention using the discovered negation direction on SimpleNeg with \textbf{ViT-B/32} , across three MLLMs-as-a-judge.}
\label{tab:negation_vitb32}
\resizebox{\textwidth}{!}{
\begin{tabular}{lccccc|ccc|cccc}
\toprule
Method &
\begin{tabular}[c]{@{}c@{}}\# Negation\\ Training Data\end{tabular} &
\begin{tabular}[c]{@{}c@{}}Without\\Finetuning? \end{tabular}
& \multicolumn{3}{c}{\textbf{Qwen3-VL-32B}} & \multicolumn{3}{c}{\textbf{Qwen3-VL-8B}} & \multicolumn{3}{c}{\textbf{Qwen3-VL-4B}} \\
\cmidrule(lr){4-6} \cmidrule(lr){7-9} \cmidrule(lr){10-12}
& & & Top-1 & Avg.5 & Top-5 & Top-1 & Avg.5 & Top-5 & Top-1 & Avg.5 & Top-5 \\
\midrule
Baseline \cite{Radford2021LearningTV} &
-- & --
& 44.9 & 44.9 & 84.7 & 46.0 & 45.2 & 84.6 & 46.9 & 45.0 & 85.4 \\
ConCLIP \cite{Singh2025LearningTP} &
228K & \textcolor{red}{\xmark}
& 30.7 & 30.2 & 63.8 & 31.7 & 30.5 & 64.6 & 30.9 & 30.3 & 63.9 \\
NegCLIP \cite{yuksekgonul2023when} &
120K & \textcolor{red}{\xmark}
& 46.0 & 45.7 & 83.1 & 47.7 & 46.7 & 84.8 & 48.1 & 46.8 & 84.2 \\
CLIP-CC12M \cite{Alhamoud2025VisionLanguageMD} &
12M & \textcolor{red}{\xmark}
& 53.1 & 49.8 & 88.0 & \textbf{55.9} & 50.8 & 89.3 & \textbf{55.0} & 50.0 & 88.3 \\ \midrule
\textbf{Steering (Ours)} &
4K & \textcolor{DarkGreen}{\cmark}
& \textbf{54.3} & \textbf{53.3} & \textbf{98.0} & 53.7 & \textbf{51.7} & \textbf{94.0} & 54.0 & \textbf{51.5} & \textbf{94.1} \\
\bottomrule
\end{tabular}
}
\end{table*}


\noindent \textbf{Controlled evaluation with SimpleNeg:} To objectively evaluate this capability, we curated a simple, controlled dataset of negation queries containing at most two objects, adjectives, or actions, and a negation (\eg, a supermarket scene with a shopping cart but no cashier). In this example, the two objects are \{scene, shopping cart\} and the negated object is \{cashier\}. We used an LLM in combination with human-in-the-loop to create this benchmark, which we term as SimpleNeg of 900 samples. While more complex queries \cite{Alhamoud2025VisionLanguageMD} are better for testing the model, they do not allow us to isolate confounding factors in the embedding space and systematically analyze the effect of negation. Our primary goal is not to maximize task complexity, but to rigorously test whether the negation direction can be explicitly identified and manipulated within the embedding space. Therefore, we adopted a unit-test approach rather than a stress-test one. Furthermore, in practical retrieval scenarios, user queries are typically concise and focused rather than long and syntactically complex. For example, users are more likely to search for ``a black shirt with no logos" than to formulate elaborate, multi-clause descriptions. 

After training the linear classifier, its coefficients (weights) $W^l \in \mathbb{R}^d$ define a direction in the latent space. This vector is aligned with $h^l$ and points toward the direction associated with negation, since the dot product $W^l h^l$ is 1 when $h^l$ corresponds to the latent representation of a negated caption. We first isolate the magnitude of a $W^l$ from its direction by performing a unit normalization operation on $W^l$. That is, $W_{dir}^{l} = W^l / ||W^l||$. We then steer $h^l$ in the direction of negation by:

\begin{equation}
\label{eq:steering}
    h^l = (1-\alpha)h^l + \alpha W_{dir}||h^l||
\end{equation}

\noindent where $\alpha$ is a hyperparameter that controls how much to steer the representations, and $||h^l||$ is to preserve the norm of the representations after steering, preventing them from being shifted into an out-of-distribution latent. We perform this for all layers; that is $l = [1 \dots L]$ where $L$ is the total number of layers in the CLIP Text Encoder\footnote{we also tried to steer only the layers with high test accuracy, but observed no notable differences}. 

\noindent \textbf{Results on SimpleNeg:} Results are presented in \cref{tab:negation_vitb32} for 3 MLLMs-as-a-judge using the metric from \cref{sec:metric-llm-as-a-judge}. We use Qwen3-VL-32B, Qwen3-VL-8B, and Qwen3-VL-4B as judges. Results demonstrate that steering the representations with the negation direction achieves performance surpassing all baselines on the SimpleNeg dataset. This approach demonstrates that (1) there exists a negation direction in the CLIP text embedding space that can be used to geometrically steer CLIP so that it captures and understands negation; (2) this can be achieved without any fine-tuning of CLIP; and (3) it requires only a relatively small dataset (4K samples) to train the linear classifier, which can be obtained completely independent of images, with any text-only LLM. Specifically, the linear classifier requires 0.03\% of the amount of data used in previous approaches \cite{Alhamoud2025VisionLanguageMD}. 

\noindent \textbf{Implementation Details:} We use the L-BFGS solver to train the binary classifier with no bias (intercept) for a maximum of 1000 iterations. We use the original CLIP model from OpenAI. For SimpleNeg, we set the database size to 25K, sourced from the COCO 2014 Validation Set \cite{Lin2014MicrosoftCC}. We use \texttt{gpt-5-mini-2025-08-07} from OpenAI as the LLM.

\begin{figure*}
    \centering
    \includegraphics[width=1\linewidth]{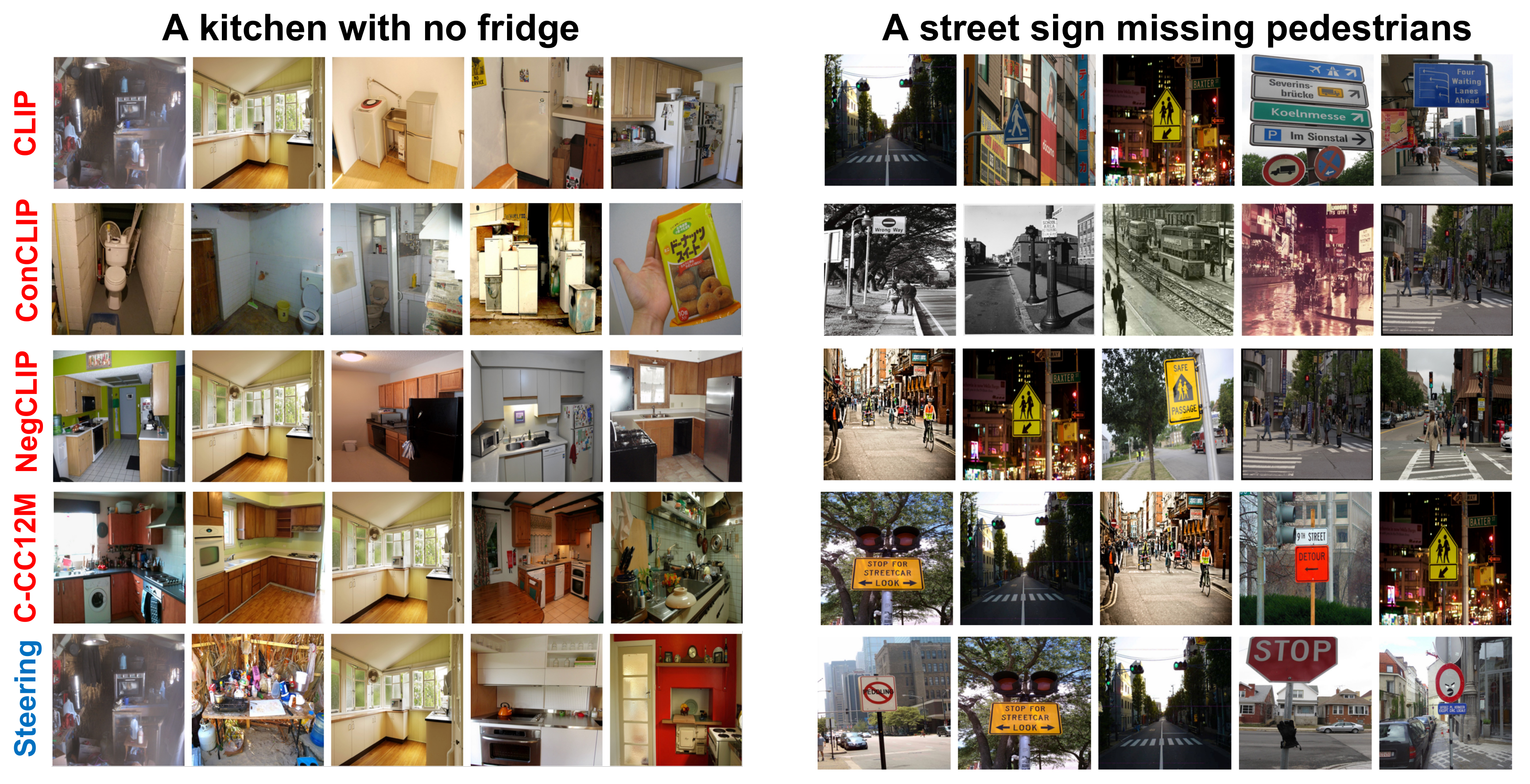}
    \caption{Qualitative examples of negated text-to-image retrieval comparing CLIP, ConCLIP, NegCLIP, CLIP-CC12M and our steering approach.}
    \label{fig:qualitative}
\end{figure*}

\begin{figure}
    \centering
    \includegraphics[width=1\linewidth]{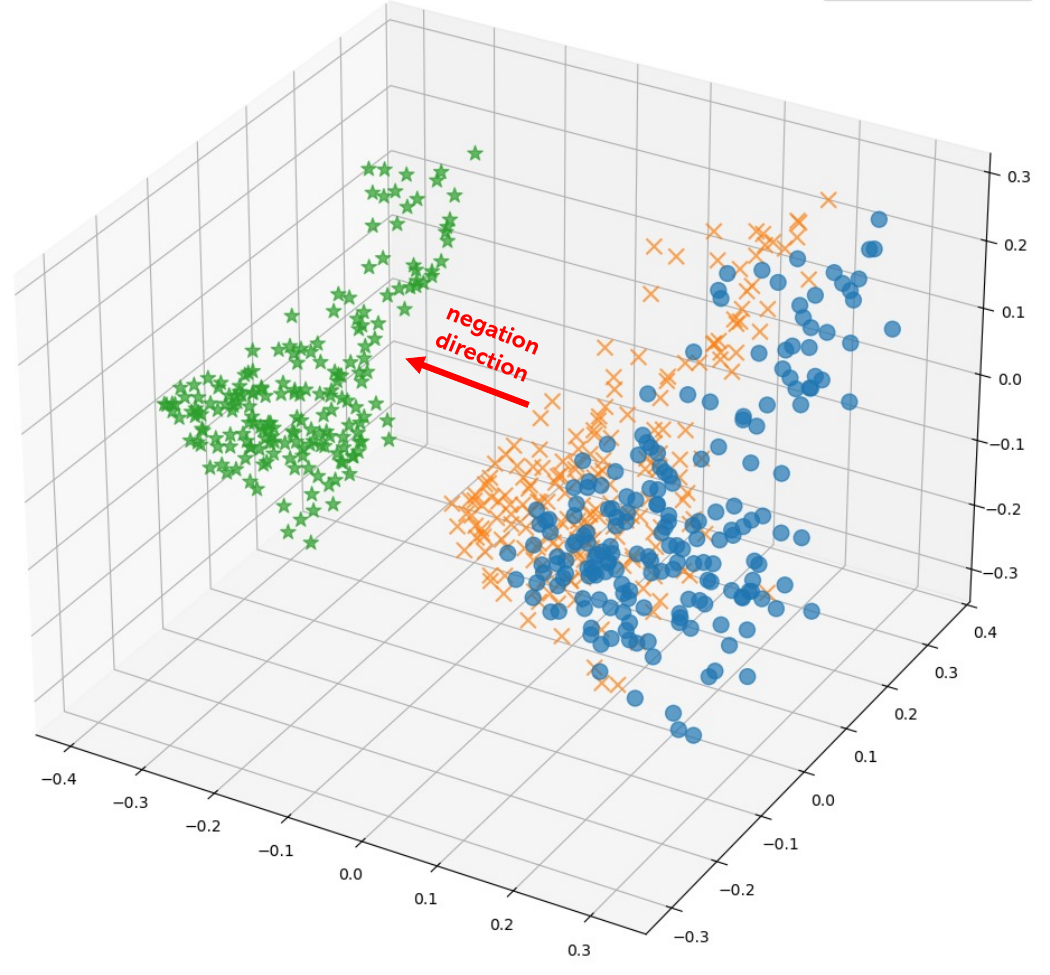}
    \caption{The {\textcolor{blue}{\(\bm{\bullet}\)}} represents the original captions, {\textcolor{orange}{\(\bm{\times}\)}} their negated counterparts, and {\textcolor{green}{\(\bm{\star}\)}} the steered representations in the negation direction.}
    \label{fig:pca}
\end{figure}

\noindent \textbf{Qualitative Examples:} We provide qualitative examples of negated text-to-image retrieval in \cref{fig:qualitative}, comparing CLIP \cite{Radford2021LearningTV}, ConCLIP \cite{Singh2025LearningTP}, NegCLIP \cite{yuksekgonul2023when}, CLIP-CC12M \cite{Alhamoud2025VisionLanguageMD} and our steering approach. Taking the first query as an example, CLIP retrieves the correct images within the top two results but fails on the last three. ConCLIP performs poorly, returning irrelevant images of a toilet. NegCLIP also fails on this query, retrieving images of a kitchen with a fridge. CLIP-CC12M makes only one incorrect retrieval, whereas the steering approach successfully retrieves the correct images in all top five.  
\newline
\newline
\noindent \textbf{Embedding Analysis:} We also visualize the embeddings of the captions in \cref{fig:pca} using Principal Component Analysis (PCA) on 197 samples. We visualize the original (affirmative) captions, their negated counterparts, and the negated embeddings after steering. As shown, the affirmative and negated representations remain somewhat separable even in a simple 3D space, supporting our hypothesis in \cref{sec:negation_dir}. However, additional refinement is still required to achieve a complete separation from the affirmative representations.

\begin{figure*}[ht]
    \centering
    \includegraphics[width=1\linewidth]{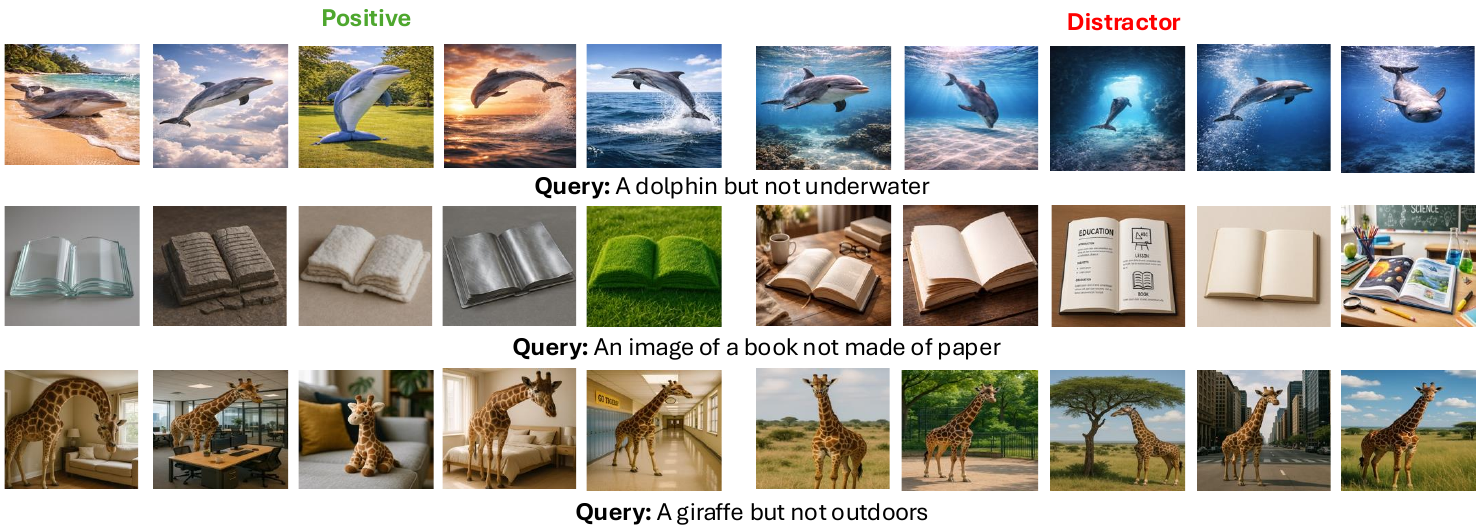}
    \caption{Visualizations of the non-common object synthetic benchmark. Positive images satisfy the negated query, while distractor images violate the negated constraint and test negation understanding and out-of-distribution generalization.}
    \label{fig:complex_benchmark_examples}
\end{figure*}

\begin{figure}
    \centering
    \includegraphics[width=\linewidth]{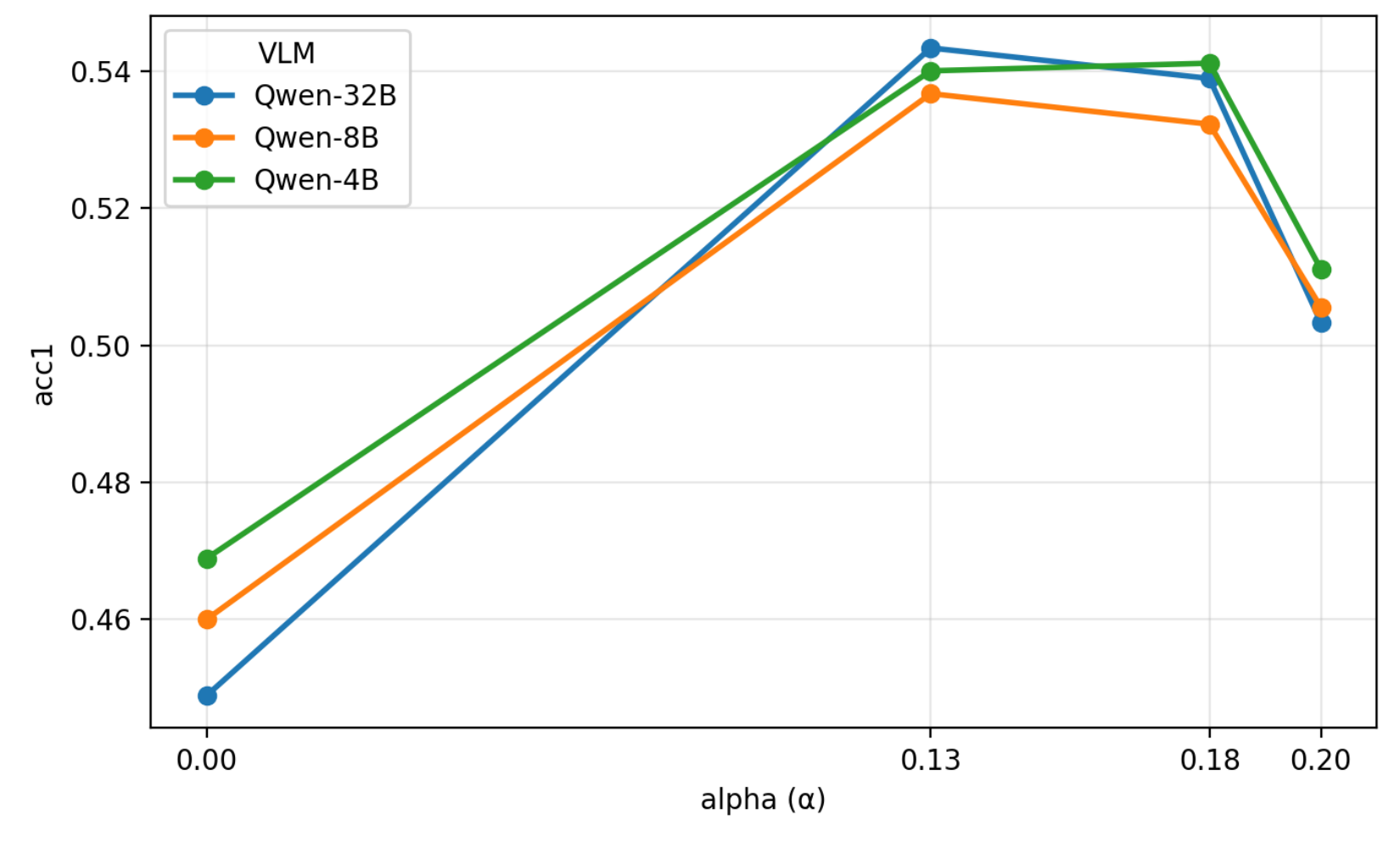}
    \caption{Ablation studies on the $\alpha$ parameter used to steer the representations. The y-axis represents the Top-1 accuracy.}
    \label{fig:ablations}
\end{figure}

\noindent \textbf{Ablation on Steering Strength:} Next, we conduct ablation studies on the $\alpha$ parameter used in \cref{eq:steering}. The results are shown in \cref{fig:ablations}. The y-axis represents the Top-1 accuracy. We observe that larger values of $\alpha$ shift the distribution excessively, leading to a degradation in performance. This behavior highlights the trade-off between amplifying the negation signal and preserving the original semantic structure of the embedding, as overly aggressive steering can distort similarity relationships. The best-performing value we identified is $\alpha = 0.13$, which we use in our experiments. This finding is consistent across all MLLM judges. 
\newline

\noindent \textbf{Non-Common Objects in Context (N-COCO):}
We also evaluate how well the baselines and our steering method generalize to non-common image–text pairs. To this end, we construct a synthetic benchmark termed as N-COCO, of 200 images spanning 10 negated queries describing uncommon scenes (\eg, ``a book not made of paper'', ``a giraffe but not outdoors''). Images are generated using GPT-5 Image with a human-in-the-loop; every generated image is manually reviewed by a human. For each query, the benchmark includes 10 positive images that satisfy the negated caption and 10 distractor images that violate it. Positive samples are produced by prompting the image generator to create plausible alternatives that meet the negation (\eg, generating a glass book for the query ``a book not made of paper"). This benchmark induces distribution shifts that are not captured by standard datasets such as COCO. An example of three queries is shown in \cref{fig:complex_benchmark_examples}.  Moreover, unlike traditional retrieval benchmarks that assume a single ground-truth image per query, our setup includes multiple semantically valid matches. This allows us to evaluate negation understanding under semantic ambiguity, where more than one image can correctly satisfy the negated condition. R@1 measures whether the top-retrieved image is one of the 10 positive ground-truth images. R@3 checks whether \textit{any} of the top 3 retrieved images are among those 10 positives, and R@5 does the same for the top 5 retrieved images. This metric overcomes the limitations of standard single-ground-truth retrieval evaluation in the presence of multiple valid matches.

\begin{table}[h]
\centering
\caption{Retrieval performance (Recall@K) of baseline models with \textbf{ViT-B/32} on the non-common objects controlled benchmark.}
\setlength{\tabcolsep}{6pt}
\begin{tabular}{lccc}
\toprule
Model & R@1 & R@2 & R@3 \\
\midrule
CLIP \cite{Radford2021LearningTV}           & 0.60 & 0.60 & 0.90 \\
NegCLIP \cite{yuksekgonul2023when}          & 0.50 & 0.50 & 0.80 \\
CLIP-CC12M \cite{Alhamoud2025VisionLanguageMD}          & 0.50 & 0.60 & 0.90 \\
Steering (ours) & \textbf{0.80} & \textbf{0.80} & \textbf{0.90} \\
\bottomrule
\end{tabular}
\par\vspace{2pt}

{\footnotesize *ConCLIP achieved 0.00 on all Recall@K metrics}
\label{tab:complex_Retrieval}
\end{table}

\begin{figure*}
    \centering
    \includegraphics[width=1\linewidth]{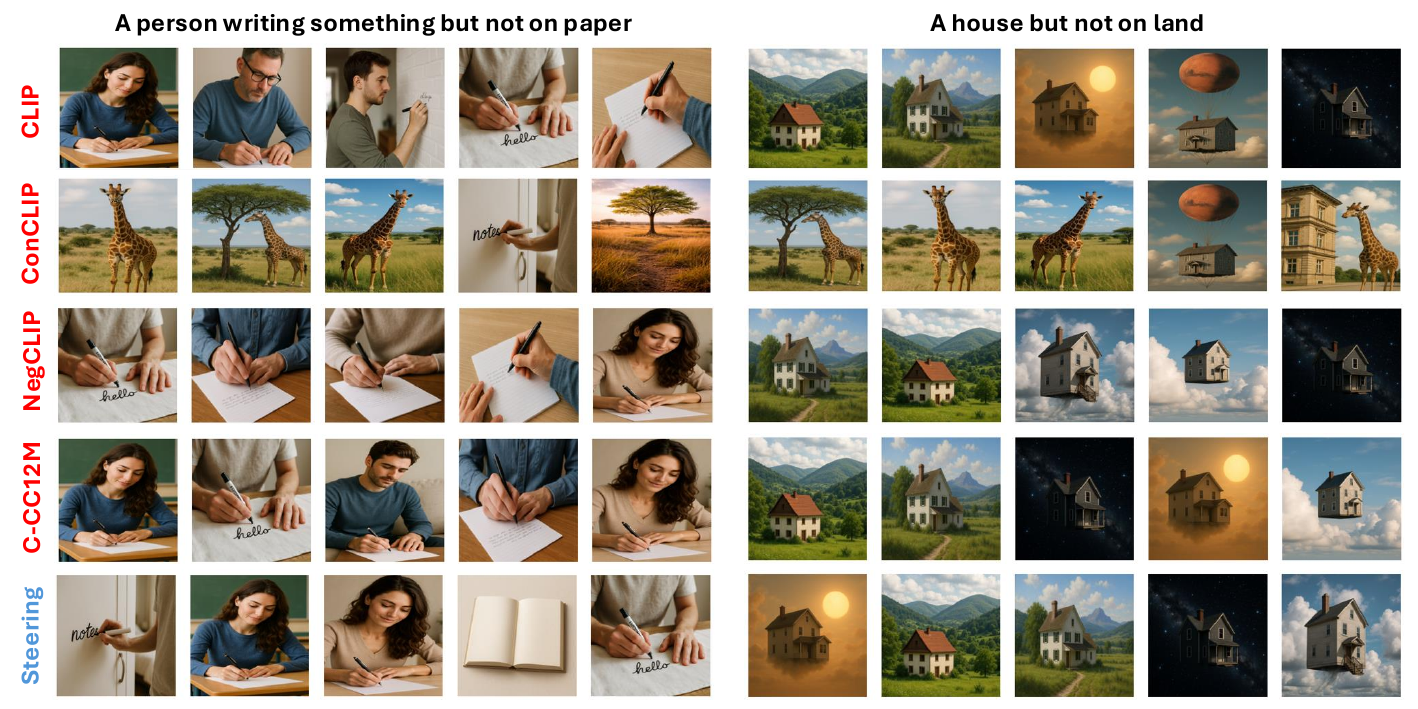}
    \caption{Qualitative examples of negated text-to-image retrieval comparing CLIP, ConCLIP NegCLIP, CLIP-CC12M and our steering approach on the N-COCO benchmark.}
    \label{fig:qualitative_complex}
\end{figure*}

\noindent \textbf{Results on N-COCO:} We report results on the N-COCO controlled benchmark in \cref{tab:complex_Retrieval}. This benchmark introduces significant distribution shift, as it contains uncommon object compositions and negated constraints not typically observed in standard benchmarks. We evaluate retrieval performance using Recall@K. As shown in \cref{tab:complex_Retrieval}, the original CLIP model maintains strong performance under this distribution shift, achieving a R@1 of 0.60. In contrast, both NegCLIP and CLIP-CC12M report a drop in performance, achieving 0.50 R@1. This is particularly notable because these models were explicitly finetuned to improve negation understanding. While prior work reports improvements on in-distribution negation benchmarks \cite{Alhamoud2025VisionLanguageMD}, our results show that this specialization does not generalize to novel compositions. The degradation suggests that finetuning on synthetic negation datasets may lead to overfitting to specific image-text patterns present in the training data, reducing robustness under distribution shift. On the contrary, our steering approach achieved the highest R@1 of 0.80. This improvement is achieved without modifying model weights or requiring additional training data.  These results support our hypothesis that negation information is already present in CLIP’s embedding space but is not consistently expressed during retrieval. Finetuning attempts to inject this capability through synthetic data but may distort the underlying geometry and reduce robustness. In contrast, representation steering operates directly on the latent structure, enabling improved negation handling without compromising generalization.

\noindent \textbf{Qualitative Examples:} Finally, we provide qualitative examples of negated text-to-image retrieval for the N-COCO benchmark in \cref{fig:qualitative_complex}, comparing CLIP \cite{Radford2021LearningTV}, NegCLIP \cite{yuksekgonul2023when}, CLIP-CC12M \cite{Alhamoud2025VisionLanguageMD}  and our steering approach. Although this benchmark is intentionally controlled, the original CLIP model maintains strong generalization, often retrieving semantically relevant images even when the object compositions are uncommon. In contrast, finetuned models such as NegCLIP and CLIP-CC12M more frequently retrieve images that violate the negated constraint. While ConCLIP fails at the retrieval task. Our steering approach produces the most consistent top-ranked results, correctly prioritizing images that satisfy both the semantic and negation requirements. 



\section{Conclusion}
In this work, we investigate negation in vision–language embedding models and identify critical flaws in existing conventional metrics that make them unreliable for assessing negation understanding. By introducing a MLLM-as-a-judge framework, we address these issues and reduce the need for data annotation, and provide a more reliable way to evaluate negation understanding in CLIP models. We then find that negation information is already encoded in CLIP’s text latent space, just not well activated. Motivated by this observation, we identified the negation direction in representations of the text encoder layers and showed that test-time steering along this direction enables negation-aware understanding. Finally, we introduced a synthetic benchmark termed as N-COCO, to evaluate CLIP’s handling of negation in uncommon scenes, and use it to assess both baseline methods and our steering approach under distribution shifts. Extending such representation-level interventions to more complex queries involving multiple interacting objects and actions remains an important direction for future work. Overall, our findings highlight the potential of lightweight representation engineering in multimodal models as an alternative to large-scale retraining.

\section*{Acknowledgment}
Fawaz Sammani is funded by the Fonds Wetenschappelijk Onderzoek (FWO) (PhD) fellowship strategic basic research 1SH7W24N). T. Chamiti and N. Deligiannis acknowledge the ”Onderzoeksprogramma Artificiele Intelligentie (AI) Vlaanderen” programme and the ERC Consolidator Grant IONIAN (No. 101171240, DOI: 10.3030/101171240). Funded by the European Union. Views and opinions expressed are however those of the author(s) only and do not necessarily reflect those of the European Union or the European Research Council Executive Agency. Neither the European Union nor the granting authority can be held responsible for them.

{
    \small
    \bibliographystyle{ieeenat_fullname}
    \bibliography{main}
}


\end{document}